\documentclass[letterpaper]{article} 
\usepackage{aaai2026}  
\usepackage{times}  
\usepackage{helvet}  
\usepackage{courier}  
\usepackage[hyphens]{url}  
\usepackage{graphicx} 
\usepackage{natbib}  
\usepackage{caption} 
\usepackage{algorithm}
\usepackage{algorithmic}
\usepackage{amsmath}
\usepackage{amsfonts}
\usepackage{booktabs}
\usepackage{xcolor}
\usepackage{url}
\newcommand{\clickurl}[1]{\pdfstartlink user{/Subtype /Link /A << /Type /Action /S /URI /URI (#1) >>}#1\pdfendlink}
\usepackage{newfloat}
\usepackage{listings}
\DeclareCaptionStyle{ruled}{labelfont=normalfont,labelsep=colon,strut=off} 
\floatstyle{ruled}
\newfloat{listing}{tb}{lst}{}
\floatname{listing}{Listing}
\title{IDCNet: Guided Video Diffusion for Metric-Consistent RGBD Scene Generation with Precise Camera Control}
\author{
    Lijuan Liu, Wenfa Li, Dongbo Zhang, Shuo Wang, Shaohui Jiao
}
\affiliations{
    Bytedance Inc. \\

    \{liulijuan, liwenfa.sjtu, zhangdongbo, wangshuo.1996, jiaoshaohui\}@bytedance.com
}

\usepackage{bibentry}

\begin{document}

\maketitle

\begin{abstract}
We present \textbf{IDC-Net (Image-Depth Consistency Network)}, a novel framework designed to generate RGB-D video sequences under explicit camera trajectory control. 
%
Unlike approaches that treat RGB and depth generation separately, IDC-Net jointly synthesizes both RGB images and corresponding depth maps within a unified geometry-aware diffusion model.
The joint learning framework strengthens spatial and geometric alignment across frames, enabling more precise camera control in the generated sequences.
%
To support the training of this camera-conditioned model and ensure high geometric fidelity, we construct a \textbf{camera-image-depth consistent dataset} with metric-aligned RGB videos, depth maps, and accurate camera poses, which provides precise geometric supervision with notably improved inter-frame geometric consistency.
%
Moreover, we introduce a geometry-aware transformer block that enables fine-grained camera control, enhancing control over the generated sequences.
%
Extensive experiments show that IDC-Net achieves improvements over state-of-the-art approaches in both visual quality and geometric consistency of generated scene sequences.
Notably, the generated RGB-D sequences can be directly feed for downstream 3D Scene reconstruction tasks without extra post-processing steps, showcasing the practical benefits of our joint learning framework.
See more at \clickurl{https://idcnet-scene.github.io}.
\end{abstract}


\section{Introduction}
\label{sec:intro}
High-fidelity, navigable 3D scenes have become the cornerstone of a wide range of applications, 
such as video gaming ~\cite{short2017procedural}, film production~\cite{anantrasirichai2022artificial}, and robotic simulation~\cite{black2024pi0visionlanguageactionflowmodel}.
Over the past decade, considerable efforts have been devoted to this area from various perspectives, such as procedural modeling~\cite{parish2001procedural}, scene layout planning~\cite{feng2023layoutgpt}, and reconstruction methods like NeRFs \cite{mildenhall2021nerf}. 
Despite their success, these approaches either require extensive manual labor or can only produce outputs with limited viewpoint flexibility, thereby restricting their applicability in broader contexts.
Recently, data-driven generative models ~\cite{rombach2021highresolution, yang2024cogvideox, wan2025wan, wang2025seededit} have demonstrated tremendous power in image/video generation tasks and have begun to achieve remarkable results in 3D object generation \cite{poole2022dreamfusion, zhang2024clay, he2025sparseflex}.
Motivated by these advances, several studies ~\cite{chung2023luciddreamer, wang2024vistadream} explored the potential of image generation models for 3D scene construction.
These methods employ training-free iterative, multi-stage pipelines that integrate pretrained depth estimation ~\cite{bochkovskii2024depth, wang2025moge} and image inpainting methods \cite{Rombach_2022_CVPR} to gradually synthesize explorable 3D scenes.
Despite their innovative approach, these methods suffer from several limitations. The complexity and sequential nature of their pipelines incur high computation cost and lead to unstable outputs. 
Moreover, a lack of holistic understanding of the scene often results in context errors and spatial inconsistencies within the generated scenes.
Driven by these challenges, recent research has turned to novel view synthesis using video generation models as alternative paradigms for 3D scene generation.
For example, methods such as Wonderland \cite{liang2025wonderland} adopt a ControlNet-style mechanism to incorporate camera trajectories into video diffusion models, enabling controllable 3D scene generations. 
Another line of work projects \cite{yu2024viewcrafter, chen2025flexworld} the input RGB-D data into multiple viewpoints to obtain partial observations, which are then completed by the video inpainting model, to facilitate novel view synthesis under the trajectory constraints. 
As observed, these models rely on either explicit or implicit integration of additional monocular depth estimation models during training and inference.
Existing monocular depth estimation models suffer from inaccurate predictions within individual frames and geometric inconsistencies across frames, as shown in Fig.~\ref{fig:data_depth}, which can mislead training and reduce the geometric fidelity of generated sequences.
Furthermore, since these models generally output only RGB image sequences without aligned depth maps, they require additional 3D reconstruction techniques, such as pose estimation and depth estimation, to produce fully navigable 3D scenes.

\begin{figure*}[t]
\centering
\includegraphics[width=\textwidth]{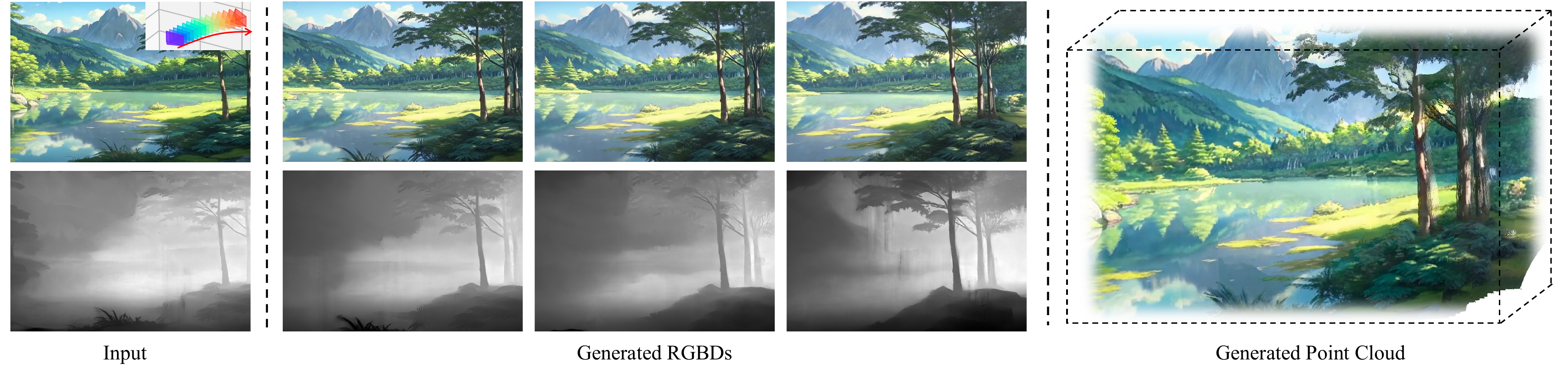}
\caption{We present \textbf{IDC-Net}, a novel framework that, given metric-aligned RGB-D inputs and camera trajectories, generates RGB-D sequences with precise camera control. Thanks to the metric alignment between generated depths and camera poses, the outputs enable direct 3D reconstruction without post-processing.}
\label{fig:teaser}
\vspace{-0.2cm}
\end{figure*}

To address these challenges, we introduce \textbf{IDC-Net}, a novel framework for generating RGB-D video sequences conditioned on explicit camera trajectories. 
Unlike prior methods, IDC-Net jointly learns RGB and depth generation within a unified geometry-aware diffusion model.
This joint learning promotes enhanced spatial and geometric representations, enabling more accurate camera control and consistent scene outputs (Fig~\ref{fig:teaser}).
A key component of our approach is the construction of a carefully curated dataset featuring metric-aligned RGB videos, depth maps, and accurate camera poses.
This dataset provides precise geometric supervision, helping to alleviate the inaccuracies and temporal inconsistencies commonly encountered when using external monocular depth estimators.
To further improve camera-conditioned generation and spatial coherence, we introduce a geometry-aware transformer block that incorporates token-level camera pose information and enhances depth awareness within the diffusion model, while maintaining architectural simplicity.
We also adopt a progressive training strategy: the model is first trained to predict depth sequences from RGB sequences, and subsequently finetuned to joint RGB-D generation under camera control.
In summary, we make the following contributions:

\begin{itemize}
    \item We propose IDC-Net, a novel framework for jointly generating RGB-D video sequences conditioned on explicit camera trajectories. This joint learning approach improves spatial and geometric consistency, resulting in more accurate and controllable 3D scene videos.
    
    \item We introduce an efficient dataset construction methodology that ensures metric alignment and consistent camera poses, providing reliable geometric supervision to train our camera-conditioned generative model.
    
    \item We perform extensive quantitative and qualitative experiments demonstrating that IDC-Net achieves superior camera control and geometric coherence in generated RGB-D sequences. Furthermore, the output can be directly applied to downstream 3D reconstruction tasks without requiring additional post-processing, validating the practical utility of our approach.
\end{itemize}

\section{Related Work}
\label{sec:rwork}

\subsection{Generative approaches for 3D synthesis} 
Early research on 3D scene generation from single image inputs relied mainly on image diffusion models ~\cite{rombach2022high, huang2025diffusion} combined with depth prediction models~\cite{yang2024depth, rohan2025systematic}, adhering to an iterative frame-by-frame incremental generation pipeline.
For instance, Luciddreamer \cite{chung2023luciddreamer}  and RealmDreamer \cite{shriram2024realmdreamer} construct 3D scenes by directly leveraging pretrained image inpainting and depth estimation models, without further fine-tuning. 
After each incremental step of inpainting and depth estimation, the 3D scene - represented in the 3DGS format~\cite{kerbl3Dgaussians} - is optimized according to the inpainted region.
RGBD2 \cite{lei2023rgbd2} shortens this pipeline by introducing a self-tuned RGB-D inpainting model specifically designed for indoor scene generation.
VistaDream \cite{wang2024vistadream} firstly employs the above-described iterative pipeline to generate an initial scene, which is then refined through a training-free multi-view consistency sampling process to enhance the coherence of the generated scenes. 
%
%
WonderWorld \cite{yu2025wonderworld} and Wonderturbo \cite{ni2025wonderturbo} further accelerates this pipeline by incorporating improved inpainting, depth estimation and gaussian splatting modules.
Although frame-by-frame scene generation methods can produce visually compelling results, the absence of holistic scene understanding often introduces semantic discrepancies and spatial inconsistencies within the generated scenes.

\subsection{Camera-Controlled Video Synthesis for 3D Scene.}
In recent years, camera-controlled scene generation has become a focal point of significant interest~\cite{lei2025animateanything, bai2025recammaster, bahmani2025ac3d, zheng2025vidcraft3, shuai2025free}. 
Previous approaches often demanded dense input data and struggled with handling single-image input ~\cite{gao2024cat3d, wu2024reconfusion, liu2024reconx, wang2025videoscene}.
More recent efforts have focused on generating 3D scene videos from just a single image.
AnimateDiff ~\cite{guo2023animatediff} injects a limited set of fixed camera motions into the model via a learned LoRA adapter, whereas MotionCtrl ~\cite{wang2024motionctrl} allows arbitrary camera movements by embedding the camera pose matrix. 
Based on this, CameraCtrl ~\cite{he2024cameractrl} enhances camera control accuracy by representing the camera pose matrix using Pl\"ucker ray. 
To further improve control precision and stability, I2VControl-Camera ~\cite{feng2024i2vcontrol} uses point trajectories in the camera coordinate system as control signals. 
However, these methods often suffer from inconsistencies in perspective, which substantially degrade performance in downstream tasks such as scene reconstruction.
Another line of research improves 3D spatial consistency by incorporating depth information into the models.
ViewCrafter \cite{wang2024vistadream} and FlexWorld \cite{chen2025flexworld} first estimate the depth of the initial frame, then project the frame onto multiple camera viewpoints to obtain partial views. 
These partial views are used to train a video-to-video (V2V) inpainting model that enables novel view synthesis. 
Building upon this, RealCamI2V \cite{li2025realcam} maintains an interactive 3D scene during inference to provide reference information, while Gen3C \cite{ren2025gen3c} employs a 3D cache during training to assist the learning process.
However, warped partial views still retain artifacts that can adversely affect model training. Moreover, existing methods are generally limited to predicting image sequences alone. 
In this work, we tackle these challenges by jointly learning RGB and depth generation using a carefully constructed dataset aligned across cameras, images, and depth maps.

\begin{figure}[tb]
\centering
\includegraphics[width=\linewidth]{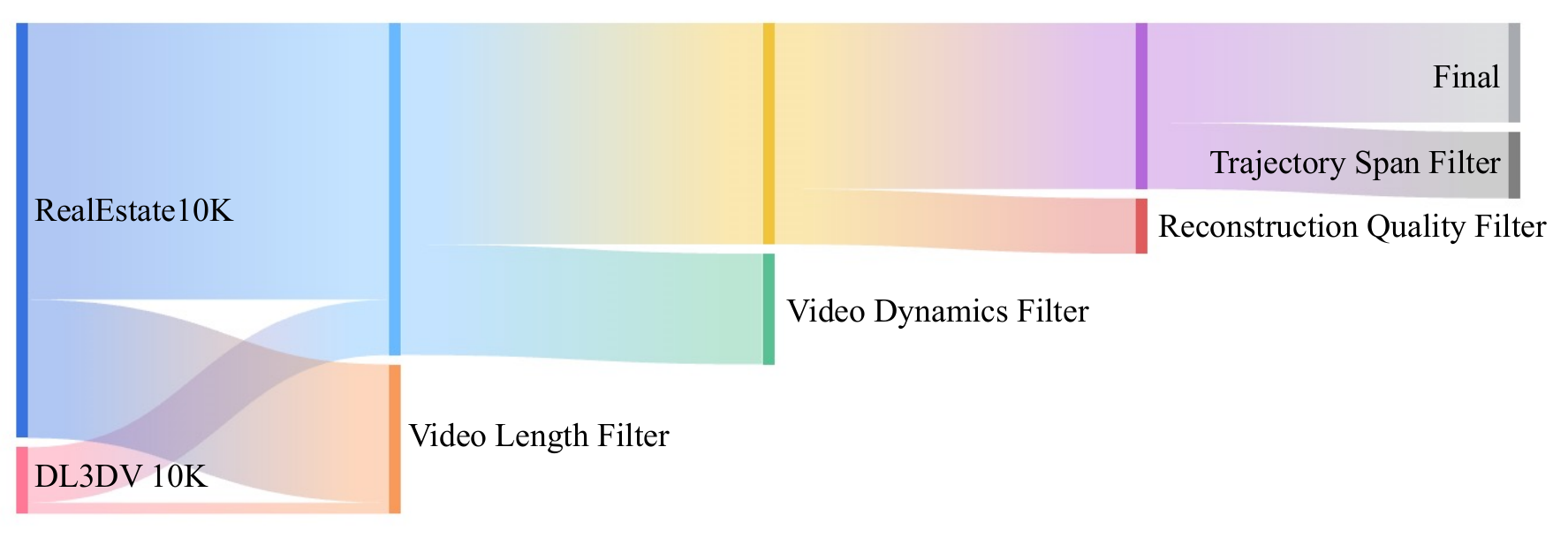} 
\caption{Overview of our data curation pipeline, which refines RealEstate10k and DL3DV-10K into about 18,000 clips for model training.}
\label{fig:data_c}
\end{figure}

\section{Method}
\label{sec:method}
In this work, we introduce IDC-Net, a novel architecture to generate RGB-D video sequences under explicit camera trajectory control. 
As shown in Fig.\ref{fig:framework}, our framework is supported by a camera-image-depth aligned dataset that provides accurate geometric supervision to learn spatially consistent and metrically accurate outputs (Section ~\ref{sec:dataset}).
Based on this foundation, a geometry-aware diffusion model is adopted that jointly synthesizes RGB and depth sequences, improving geometric consistency across frames (Section ~\ref{sec:rgbd}).
To further strengthen camera conditioning, we incorporate a geometry-aware transformer block that injects token-level pose information into the denoising process (Section ~\ref{sec:camera}). 

\subsection{Metric-Aligned RGB-D Trajectory Dataset}
\label{sec:dataset}
Accurate camera-conditioned RGB-D video generation requires reliable supervision of both geometric structure and camera pose. 
In this work, we build upon the RealEstate10K ~\cite{zhou2018stereo} and DL3DV-10K ~\cite{ling2024dl3dv} datasets and introduce a comprehensive data curation pipeline to obtain high-quality scene videos. 
Additionally, we develop a robust annotation pipeline to obtain metric-aligned RGB-D sequences, providing reliable geometric supervision for model training.

\begin{figure}[tb]
\centering
\includegraphics[width=\linewidth]{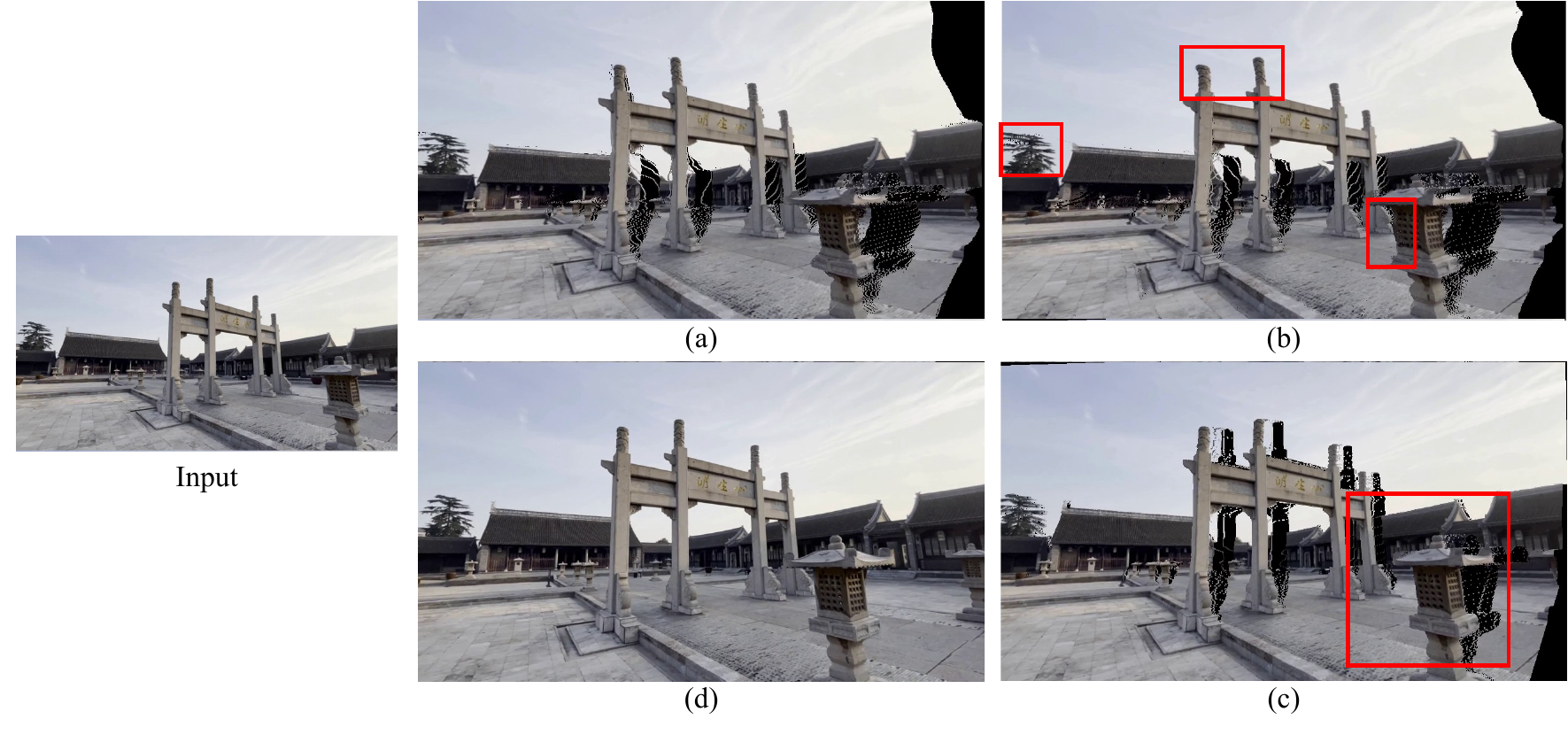} 
\caption{Depth reprojection comparison on DL3DV-10K. Given an RGB frame at input pose, we reproject the scene into a novel viewpoint using different depth annotations. Our refined depth from the proposed pipeline (a) better preserves intra-frame geometry and iter-frame consistency compared to Gaussian Splatting (b) and VideoDepthAnything aligned with COLMAP sparse depth (c). (d) is the ground truth of novel view.}
\label{fig:data_depth}
\end{figure}

\subsubsection{Data Curation.}
RealEstate10K contains 74,766 video clips, mostly depicting indoor environments, while DL3DV-10K comprises 10,000 real-world clips with more complex and rapid camera motions.
As illustrated in Fig.~\ref{fig:data_c}, to ensure the quality and relevance of training data for our training framework, we develop a four-stage data curation pipeline that filters and selects suitable clips from both datasets.

\textit{1) Video Length Filter.} To align temporal sampling, we discard RealEstate10K clips with fewer than 98 frames—ensuring a sampling stride greater than 1—and DL3DV-10K clips with fewer than 49 frames to meet the model’s minimum input requirement.

\textit{2) Video Dynamics Filter.} To eliminate scenes with undesirable motion, we use CogVLM2-Caption~\cite{hong2024cogvlm2} to generate video descriptions and exclude clips indicating dynamic content or scene changes, retaining only temporally stable scenes.

\textit{3) Reconstruction Quality Filter.} Since DL3DV-10K includes COLMAP-based point clouds, we build Gaussian Splatting models for each scene and remove clips with test-time PSNR below 24, ensuring strong geometric consistency.

\textit{4) Trajectory Span Filter.} We define a trajectory span metric that quantifies camera motion:
\begin{equation}
\mathrm{Score} = \sum_{i=1}^{N} |\Delta \mathbf{t}_i| + \sum_{i=1}^{N} |\Delta \boldsymbol{\theta}_i| \cdot \gamma,
\label{eq:trajectory-score}
\end{equation}

\begin{figure*}[t]
\centering
\includegraphics[width=0.99\textwidth]{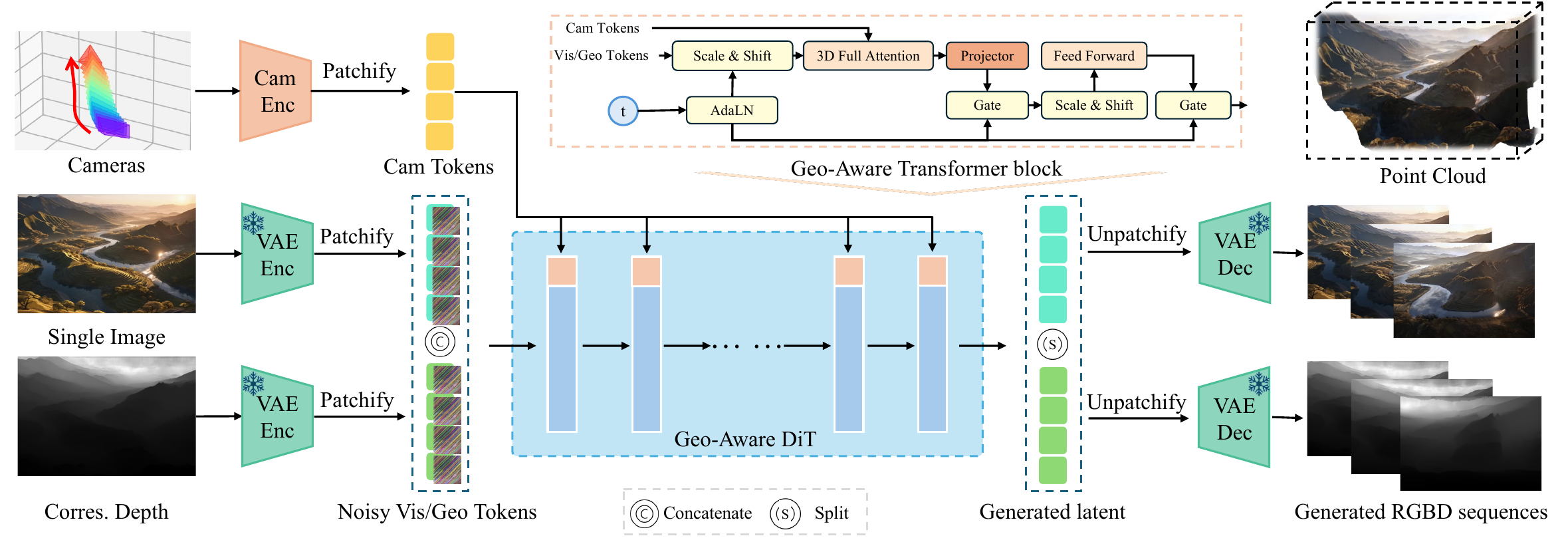} 
\caption{Overview of the IDC-Net framework. The model jointly generates RGB and depth sequences in latent space, conditioned on an input frame and target camera trajectory. Camera poses are embedded through a GeoAware transformer to enforce spatial consistency. The generated metrically aligned RGB-D outputs enable direct point cloud extraction, supporting immediate downstream 3D reconstruction (The point cloud showed above is generated based on the model outputs according to Eq. (2)).}
\label{fig:framework}
\end{figure*}

where $|\Delta t_i|$ is the translation and $|\Delta \theta_i|$ is the rotation between consecutive frames. $\gamma$ is used to balance the translation and rotation metric results. We retain RealEstate10K clips with $S$ above a threshold, and DL3DV-10K clips below it, balancing motion complexity across datasets and avoiding extreme trajectories.

\textit{Final Dataset.} This pipeline yields a curated dataset of roughly 18,000 video clips, each containing $100 \sim 300$ frames, well-suited for RGB-D video generation under explicit camera control.

\paragraph{Data Annotation.}
Accurate and metric-aligned camera poses and depth maps are essential for effective model training. 
However, the RealEstate10K and DL3DV-10K datasets do not provide such ground-truth annotations. 
To address this limitation, we design a \textit{coarse-to-fine} annotation pipeline that ensures inter-frame geometric consistency and intra-frame depth fidelity. 
In the \textit{coarse} stage, we focus on establishing global metric alignment and consistent camera trajectories. 
For DL3DV-10K, we leverage the COLMAP-generated point clouds and reconstruct depth maps using Gaussian Splatting~\cite{kerbl3Dgaussians}.
For RealEstate10K, where such reconstructions are unavailable, we apply VGGT~\cite{wang2025vggt} to jointly estimate camera parameters and coarse depth maps, providing consistent, metric-aligned camera poses and depths for multi-frame supervision.
In the \textit{refinement} stage, we improve depth quality using PriorDA ~\cite{wang2025depthprior}.
Given a RGB frame and its coarse metric depth map as prior input, PriorDA first estimates a pixel-wise alignment field that geometrically aligns the monocular depth predictions with metric prior, correcting spatial scale discrepancies.
Then, a conditioned monocular depth estimator is employed to suppress the noise and artifacts introduced by the coarse prior, recovering fine-grained and geometrically faithful depth maps.

To illustrate the effectiveness of our annotation pipeline, we visualize a depth reprojection example from the DL3DV dataset in Fig.~\ref{fig:data_depth}.
%
Given a single RGB frame at input pose and its estimated depth map $D$, we re-project the scene into a novel viewpoint
using camera intrinsics $K$ and input/output extrinics $[R, t], [R', t']$. Each pixel $\mathbf{p} = (u, v)^{\top}$ is first back-projected into 3D:
\begin{equation}
    \mathbf{X} = R^\top (D(\mathbf{p}) \cdot K^{-1} 
    \begin{bmatrix}
    u \\
    v \\
    1
    \end{bmatrix}
- t ),
\label{eq:reproj}
\end{equation}
then transformed to the novel view and projected to the image plane:
\begin{equation}
    \mathbf{p}' = K \cdot \frac{R' \cdot \mathbf{X} + t'}{Z'}
\label{eq:proj}
\end{equation}
where $Z'$ is the depth of the transformed point.
As shown, our pipeline produces depth maps with high intra-frame geometric fidelity and accurate inter-frame consistency. 
Compared to depths derived from Gaussian Splatting reconstruction (Fig.~\ref{fig:data_depth}(b)), our refined outputs (Fig.~\ref{fig:data_depth}(a)) better preserve fine-grained scene structures. 
Moreover in Fig.~\ref{fig:data_depth}, our method achieves superior geometric coherence than VideoDepthAnything~\cite{chen2025video} aligned with COLMAP sparse depth (Fig.~\ref{fig:data_depth}(c)), highlighting the advantage of our coarse-to-fine annotation strategy for generating metrically consistent RGB-D sequences.

\subsection{Joint RGBD Video Diffusion}
\label{sec:rgbd}
Our RGB-D video generation framework, \textbf{IDC-Net}, builds upon the CogvideoX~\cite{yang2024cogvideox} architecture, a latent video diffusion model that operates in a compressed representation space.
We extend this framework to simultaneously model both RGB appearance and geometric depth signals under explicit camera trajectory control.
By integrating depth generation directly into the diffusion process, IDC-Net leverages geometric cues during training, resulting in stronger spatial alignment and improved geometric consistency across frames.
This joint learning facilitates more accurate camera control and enables the synthesis of temporally and structurally consistent RGB-D video sequences.
An overview of the proposed framework is illustrated in Fig.~\ref{fig:framework}.

\begin{figure*}[ht!]
\centering
\includegraphics[width=0.9\textwidth]{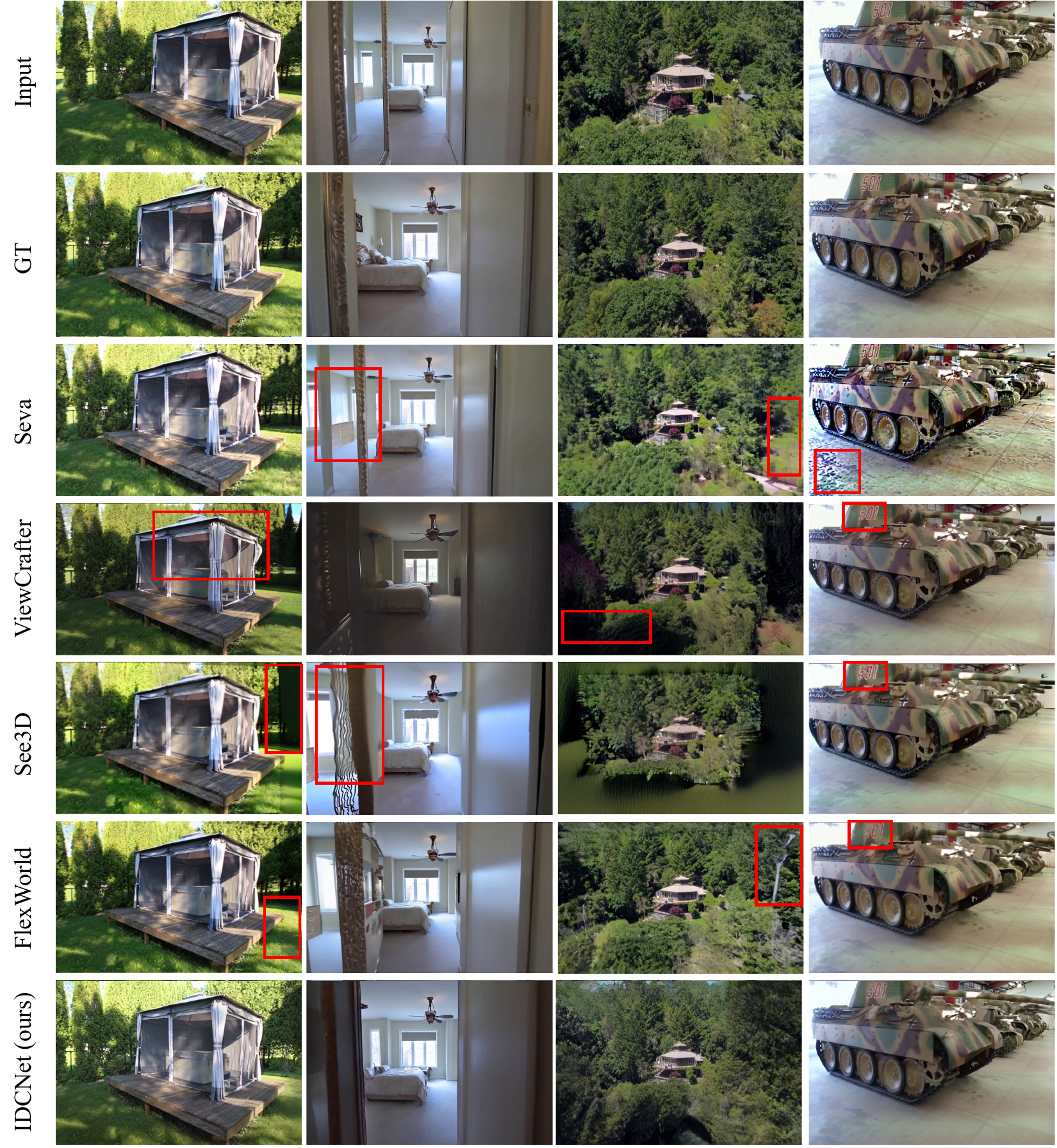} 
\caption{Qualitative results of camera-controllable video generation. Left three cols: RealEstate; Right: Tanks-and-Temples. Compared to prior methods, IDC-Net better preserves known scene content while generating semantically consistent and coherent novel views under target camera trajectories.}
\label{fig:sota}
\end{figure*}

\paragraph{Latent Video Diffusion.} 
IDC-Net adopts a latent video diffusion paradigm, CogVideoX ~\cite{yang2024cogvideox}, to generating RGB-D sequences in a compressed representation space.
Specifically, input RGB video frames are first encoded into compact latent representations $\mathbf{z}_0$ using a pretrained Variational Autoencoder (VAE).
%
A forward diffusion process then gradually adds Gaussian noise over $T$ timesteps, according to:
\begin{equation}
\label{eq:01}
q(\mathbf{z}_t | \mathbf{z}_{t-1}) = \mathcal{N}(\mathbf{z}_t; \sqrt{1 - \beta_t} \mathbf{z}_{t-1}, \beta_t \mathbf{I})
\end{equation}

where $ t \in \{1, 2, \dots, T\} $ denotes the diffusion step, $\{\beta_t\}_{t=1}^{T}$ is a predefined noise schedule. $ \alpha_t = 1 - \beta_t $ control noise magnitude at each step, and $ \mathbf{I} $ is the identity matrix. 
During the reverse process, the model predicts the noise component $\epsilon_\theta(\cdot)$ to iteratively denoise $\mathbf{z}_t$ and estimate the clean latent $\mathbf{z}_0$. A single reverse update step is given by:
\begin{equation}
\label{eq:02}
\mathbf{z}_{t-1} = \sqrt{\alpha_{t-1}} \frac{\mathbf{z}_t - \sqrt{1-\alpha_t} \epsilon_\theta(\mathbf{z}_t, t)}{\sqrt{\alpha_t}} + \sqrt{1-\alpha_{t-1}} \epsilon_\theta(\mathbf{z}_t, t).
\end{equation}

The denoising network $\epsilon_\theta$ is trained to minimize the simplified loss of noise estimation:
\begin{equation}
\label{eq:03}
\mathcal{L}(\theta) = \mathbb{E}_{z_0, \epsilon, t} \big| \epsilon - \epsilon\theta(z_t, t) \big|^2,
\end{equation}
where $\epsilon \sim \mathcal{N}(0, \mathbf{I})$ is Gaussian noise and $\mathbf{z}_t = \sqrt{\alpha_t} \mathbf{z}_0 + \sqrt{1-\alpha_t} \epsilon$.  
Using the trained diffusion model, the latent representation $\mathbf{z}_0$ is denoised and subsequently decoded back into pixel space to produce the final RGB video via the CogVideoX VAE decoder.

\paragraph{Joint RGB-D Modeling.} To improve geometric consistency in video generation, 
we adopt a joint modeling approach for RGB and depth sequences under explicit camera trajectory condition.
Given an input image $I$, its corresponding depth map $D$, and a target camera trajectory $\mathcal{T}=\{p_0, \dots, p_N\}$, IDCNet is trained to generate the corresponding RGB-D video sequence $(\mathcal{V}, \mathcal{D})$ simultaneously.
%
Formally, the model learns the conditional distribution $p(v, d|v_0, d_0, \mathcal{T})$ , where $v \in \mathbb{R}^{c\times f \times h \times w}$ and $d \in \mathbb{R}^{c\times f \times h \times w}$ denote the latent representations of the RGB and depth video sequences, respectively. 
The conditional latent embeddings $(v_0, d_0)$ are obtained by encoding the input image and depth map using a shared VAE encoder:
\begin{equation}
(v_0, d_0) = \mathrm{Enc}_{\mathrm{VAE}}(I, D).
\label{eq:vae-encode}
\end{equation}
For efficient joint processing, we concatenate $v_0$ and $d_0$, forming a unified latent tensor $\mathbf{x}_0=[v_0;d_0]$, which serves as input to video diffusion model. 
During training, the model is optimized to denoise the noisy inputs under trajectory conditioning $\mathcal{T}$, guided by the loss in Eq~\eqref{eq:03}.

%
After denoising, the output latent tensor $\hat{\mathbf{x}}_0$ is split back into individual RGB and depth components $(\hat{v}_0$, $\hat{d}_0$), which are then decoded independently into the pixel space:

\begin{equation}
(\hat{\mathcal{V}}, \hat{\mathcal{D}}) = \mathrm{Dec}_{\mathrm{VAE}}(\hat{v}_0, \hat{d}_0),
\label{eq:vae-decode}
\end{equation}

where $\hat{\mathcal{V}}$ and $\hat{\mathcal{D}}$ represent the final RGB and depth video frames, respectively. 
%
This joint latent formulation encourages the model to learn stronger cross-model correlations and enforce geometric consistency across frames, which significantly enhances spatial alignment under varying camera trajectories.

\subsection{Geo-Aware Camera Token Injection}
\label{sec:camera}
To enable precise control over camera motion in RGB-D video generation, we inject camera pose information into the denoising process using a geometry-aware token injection mechanism (see Fig.~\ref{fig:framework}). 
Inspired by CameraCtrl ~\cite{he2024cameractrl}, we represent each target pose in the trajectory $\mathcal{T}=\{p_0, p_1, \dots, p_N\}$, where $p_i \in \mathrm{SE}(3)$, via a dense field of Plücker rays computed over the image grid.

Each camera pose $p_i = [\mathbf{R}_i|\mathbf{t}_i]$ is used to project image-plane coordinates through the inverse intrinsic matrix $\mathbf{K}^{-1}$, yielding direction vectors $\mathbf{d}^{(u, v)}$ for each pixel $(u, v)$. 
The Plücker ray at each pixel is then constructed as: 
\begin{equation}
    \mathbf{l}^{(u, v)} = \begin{bmatrix}  \mathbf{d}^{(u,v)};  \mathbf{o}^{(u,v)} \times \mathbf{d}^{(u,v)} \end{bmatrix},
\end{equation}
where $\mathbf{o}^{(u,v)}$ is the camera origin. 
This results in a dense Plücker field $\mathbf{L}_i \in \mathbb{R}^{6\times h \times w}$, which is then embedded via a lightweight encoder and patchified into a token sequences $\mathbf{c}_i\in \mathbb{R}^{n_c \times d}$, aligned with the latent visual tokens.
We integrate these camera tokens into each Transformer layer of the denoising network by slightly modifying the attention computation  as:
\begin{equation}
    \mathrm{Attn}(\mathbf{z}_t) \gets \mathrm{SelfAttn}(\mathbf{z}_t) + \mathrm{CrossAttn}(\mathbf{z}_t, \mathbf{c}),
\end{equation}
where $\mathbf{z}_t \in \mathbb{R}^{n_f\times d}$ denotes the latent feature tokens at timestep $t$, and $\mathbf{c}\in \mathbb{R}^{n_c\times d}$ is the set of Plücker-based camera tokens.
The self-attention operates on $\mathbf{z}_t$ as usual, while the cross-attention allows latent queries to attend to geometric context encoded in $\mathbf{c}$. Formally, this cross-attention is implemented as:
\begin{equation}
     \mathrm{CrossAttn}(\mathbf{z}_t, \mathbf{c}) = \mathrm{softmax}(\frac{\mathbf{Q}\mathbf{K}_\mathbf{c}^{\top}}{\sqrt{d}})\mathbf{V}_\mathbf{c}
\end{equation}
where $\mathbf{Q}=\mathbf{z}_t\mathbf{W}^Q, \mathbf{K}=\mathbf{c}\mathbf{W}^K, $ and$ \mathbf{V}=\mathbf{c}\mathbf{W}^V$ are linear projections of the queries and camera tokens. 
These terms are fused into the latent representation to ensure that pose-aware geometric context directly modulates the generation process.
To improve compatibility with the pre-trained CogVideoX backbone, we introduce a projection layer that aligns the geometry-enriched latent space with the model's original token distribution. 
This alignment improves the training stability and promotes geometric consistency in both RGB and depth outputs.

\begin{figure*}[t]
\centering
\includegraphics[width=0.99\textwidth]{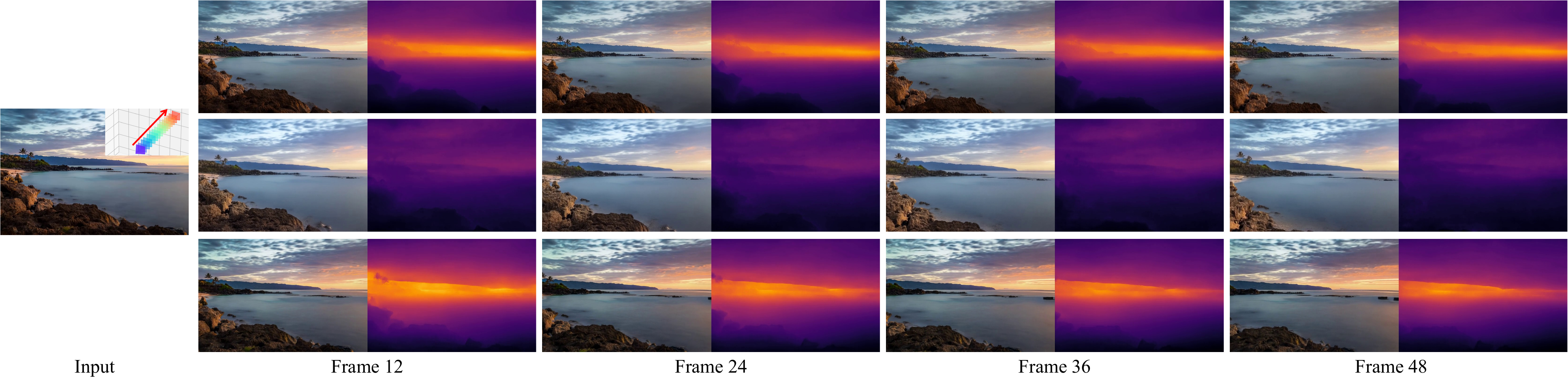} 
\caption{Qualitative ablation results of IDC-Net.
Given the same input image and predefined trajectory (“Move Forward”), the full IDC-Net model produces visually coherent videos with accurate depth and trajectory adherence (Top Raw). In contrast, the one-step baseline introduces noticeable noise in depth prediction (Middle Raw), while the IDC-Net-Var variant fails to follow the intended camera motion (Last Raw) (please refer to the supplementary material for dynamic visualizations).}
\label{fig:ab}
\end{figure*}

\begin{table}[t]
    \caption{Quantitative comparison against prior art in camera guided scene video generation on RealEstate10K. (\textbf{Bold} and \underline{underline} indicate the best and second-best results, respectively.)
    }
    \resizebox{\linewidth}{!}{
        \centering
        \begin{tabular}{lccccc}
        \toprule
        Method & PSNR $\uparrow$ & SSIM $\uparrow$ & LPIPS $\downarrow$  &  $R_{\text{err}}$ $\downarrow$ & $T_{\text{err}}$ $\downarrow$\\
        \midrule
        Seva & \underline{15.49} & 0.523 & 0.389 & 0.030 & \textbf{0.085}\\
        \midrule
        ViewCrafter & 15.42 & 0.533 & 0.396 &  0.086 & 0.153  \\
        See3D & 15.32 & \underline{0.534}   &  \underline{0.381} &  \textbf{0.015} & 0.098 \\
        FlexWorld & 14.62 & 0.496  & 0.395 & \underline{0.022} & 0.105 \\
        \midrule
        IDC-Net & \textbf{16.72} & \textbf{0.567} & \textbf{0.345} & \underline{0.022} & \underline{0.092} \\
        
        \bottomrule
        \end{tabular}
        }

\label{tab:sota-full-1}
\end{table}

\section{Experiments}
\subsection{Implementation Details}
%
We train IDCNet on the dataset introduced in Section~\ref{sec:dataset}. 
The training procedure largely follows the CogVideoX-5B-I2V generation framework. 
The model is configured to generate videos consisting of 49 frames at resolution of $480\times 720$.
It utilizes the 3DVAE framework to compress video clips, applying temporal and spatial downsampling factors of $r_t=4$ and $r_s=8$, respectively, to produce a latent representation of size $13\times60\times90$.
To enable joint RGB-D generation, we input RGB and depth videos separately into a shared 3DVAE encoder, the resulting latents are concatenated along the channel dimension, forming a joint latent tensor of shape $26\times60\times90$. 
Correspondingly, the input and output layers of the diffusion model are adapted to accommodate the increased latent dimensionality.

\paragraph{Training Strategy.} 
During training, we apply different frame sampling strides: a stride of $[2, 4]$ for video clips from the RealEstate10K dataset and a stride of 1 for those from the DL3DV-10K dataset.
To effectively incorporate depth information, we adopt a two-stage training strategy. 
First, we train an RGB-to-RGBD video generation model, allowing the network to learn depth-aware representations.
Based on this pretrained model, IDC-Net is subsequently trained in the second stage. 
All adapter and projection layers responsible for camera parameter injection are zero-initialized at the beginning of this stage. 
Please refer to the supplementary material for additional training details and model architecture configurations.

\subsection{Evaluation} 
\label{sec:eval}
We evaluate the quality of controlled scene video generation by IDC-Net through comparison with four open-source, camera-controllable image-to-video generation methods: Seva~\cite{zhou2025stable}, ViewCrafter ~\cite{yu2024viewcrafter}, See3D~\cite{ma2025you}, and FlexWorld~\cite{chen2025flexworld} and .
Seva directly inputs camera representations as conditions, while last three methods incorporate explicit depth estimation to reconstruct point clouds, which serves as geometric references for controlling camera trajectories. The point clouds are respectively derived from pretrained models on DUst3R~\cite{wang2024dust3r}, MoGE~\cite{wang2025moge} and MAst3R~\cite{leroy2024grounding}.

\begin{table}[t]
    \caption{Quantitative evaluation of camera-guided scene
video generation performance on the Tanks-and-Temples
dataset. 
    }
    \resizebox{\linewidth}{!}{
        \centering
        \begin{tabular}{lccccc}
        \toprule
        Method & PSNR $\uparrow$ & SSIM $\uparrow$ & LPIPS $\downarrow$  &  $R_{\text{err}}$ $\downarrow$ & $T_{\text{err}}$ $\downarrow$\\
        \midrule
        Seva & 12.47 & 0.364 & 0.460 & 0.056 & \textbf{0.092} \\
        \midrule
        ViewCrafter & 12.68 & 0.340 & 0.514 & 0.248 & 0.261\\
        See3D & \underline{13.14} & \underline{0.375} & \underline{0.420} & \textbf{0.026} & \underline{0.103} \\
        FlexWorld &12.80 & 0.350 & 0.461 & 0.048 & 0.129 \\
        \midrule
        IDC-Net & \textbf{14.08} & \textbf{0.442} & \textbf{0.358} & \underline{0.044} & 0.191 \\
        
        \bottomrule
        \end{tabular}
        }

\label{tab:sota-full-tt}
\end{table}

\paragraph{Datasets.}
\label{sec:eval-data}
We conduct evaluation on two benchmark datasets: RealEstate10K~\cite{zhou2018stereo} and Tanks-and-Temples~\cite{Knapitsch2017}. 
%
In RealEstate10K, 150 video clips are sampled with frame strides ranging from 1 to 3, while in Tanks-and-Temples, 100 video clips are extracted with a fixed stride of 4 across 14 test scenes.
Notably, both datasets do not provide ground-truth RGB-D annotations. 
To address this, we apply our previously introduced pipeline (Section.~\ref{sec:dataset}) to annotated metric-aligned camera poses and depth maps, ensuring consistent and accurate supervision for evaluation.

\paragraph{Evaluation Metrics.}
\label{sec:eval-metric}
To comprehensively assess the generated videos, we evaluate visual quality using standard metrics including PSNR, SSIM and LPIPS, which quantify frame-wise fidelity to the ground truth.
To measure camera controllability, we estimate camera poses for both the generated videos and ground-truth sequences using VGGT~\cite{wang2025vggt}. 
Camera pose accuracy is then computed following established protocols from prior works~\cite{he2024cameractrl, chen2025flexworld, li2025realcam}, enabling quantitative evaluation of how well the generated video respects the target camera trajectory.
\paragraph{Evaluation Results.}
\label{sec:eval-res}
We perform both quantitative and qualitative evaluations to assess IDC-Net's performance on controlled scene video generation.
%
Experiment results on RealEstate10k test dataset (Table.~\ref{tab:sota-full-1}) confirm that IDC-Net delivers state-of-the-art performance in generating high-quality RGB-D video sequences and achieving robust camera control, demonstrating its superior efficacy in both visual fidelity and accurate camera trajectory adherence.
On the Tanks-and-Temples dataset (Table.~\ref{tab:sota-full-tt}), IDC-Net achieves top scores in visual quality metrics, while exhibiting slightly weaker performance in trajectory accuracy, indicating room for improvement in challenging 3D scenes.
Qualitatively, as illustrated in Fig.~\ref{fig:sota}, all baseline methods exhibit a certain degree of camera controllability but suffer from various limitations: 
Seva ~\cite{zhou2025stable} tends to produce noise artifacts in the synthesized novel views;
ViewCrafter \cite{yu2024viewcrafter} often produces darker novel views and fails to preserve known content accurately; 
%
See3D \cite{ma2025you} generates content with noticeable detail loss and conetxt inconsistency in the missing regions;
and FlexWorld \cite{chen2025flexworld} occasionally generates semantically implausible scenes. 
In contrast, IDC-Net consistently synthesizes visually coherent and temporally stable novel views, maintaining strong geometric and appearance consistency with the input. 
This improvement validates the benefits of our joint RGB-D latent modeling and geometry-aware transformer design, which together enhance spatial consistency and camera-guided generation (more video results are provided in the supplementary material).

\begin{table}[t]
    \caption{Quantitative ablation experiment results on RealEstate10K test dataset. (IDC-Net-onestep: Progressive Training Strategy; IDC-Net-var: Token-Level Camera Injection)}
    \resizebox{\linewidth}{!}{
        \centering
        \begin{tabular}{lccccc}
        \toprule
        Method & PSNR $\uparrow$ & SSIM $\uparrow$ & LPIPS $\downarrow$ &$R_{\text{err}}$ $\downarrow$ & $T_{\text{err}}$ $\downarrow$ \\
        \midrule
        IDC-Net-onestep & 14.58 & 0.516 & 0.399 & 0.028 & 0.109\\
        IDC-Net-var & 15.31 & 0.519 & 0.380 & 0.061 & 0.206\\
        \midrule
        IDC-Net & 16.72 & 0.567 & 0.345 & 0.022 & 0.092 \\
        \bottomrule
        \end{tabular}
    }
\label{tab:ab-1}
\end{table}

\subsection{Ablation Studies}
\label{eval:ab}
To systematically evaluate the efficacy of our proposed framework, we conduct ablation studies targeting two key components: the progressive training strategy and the token-level camera injection mechanism.

\begin{itemize}
    \item \textbf{Progressive training strategy.} Implemented as IDC-Net-onestep, this variant omits the RGB-to-RGBD video training phase to isolate the impact of our progressive approach, directly fine-tuning of the pre-trained CogVideo-5B-I2V checkpoint in a single stage. 
    
    \item \textbf{Token-Level Camera Injection.} Realized as IDC-Net-var, this variant injects camera tokens at the initial layer of the DiT module. ControlNet-style injection is excluded due to its reliance on a well-pretrained backbone, whereas our method injects camera representations via supervised fine-tuning with appropriate initialization.
\end{itemize}

Quantitative results are reported in Table ~\ref{tab:ab-1}.
These results indicate that the progressive training strategy enhances visual quality, while the proposed camera token injection module markedly boosts  the model’s capacity to control camera trajectories.
We additionally provide qualitative comparisons in Fig.~\ref{fig:ab}. 
Given a single input image and a predefined camera trajectory (e.g., “Move Forward”), our model is capable of generating realistic scene videos, featuring high-quality depth reconstruction and precise camera control.
In comparison, the one-step training baseline produces noisy depth outputs, and the IDC-Net-Var model fails to follow the intended trajectory, generating a video corresponding to a “Move Right” motion instead.
Collectively, these qualitative and quantitative results validate the effectiveness of our model architecture and training strategy.

\begin{figure}[tb]
\centering
\includegraphics[width=0.96\linewidth]{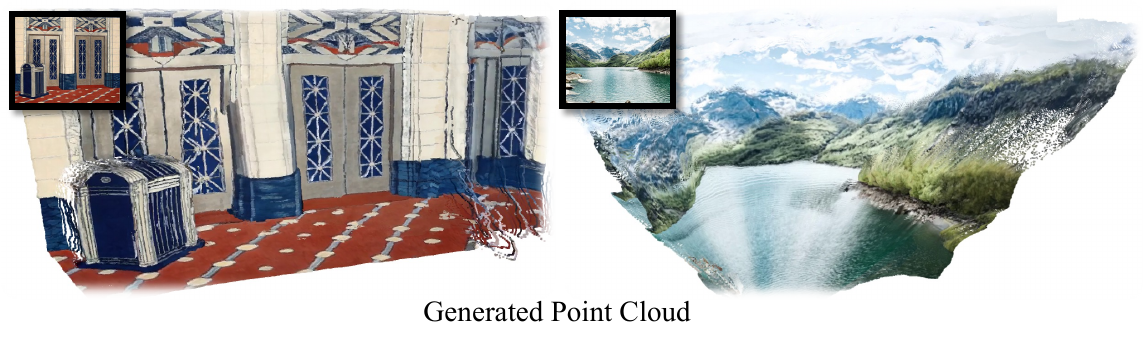} 
\caption{Examples of input images (top-left) and their corresponding metrically aligned 3D point clouds derived from IDC-Net outputs.}
\label{fig:pointcloud}
\end{figure}

\subsection{Application} 
\label{sec:app}
Our model generates RGB-D video sequences conditioned on precise camera trajectories, 
which not only improves generation fidelity and geometric consistency but also produces metrically aligned outputs due to training on a carefully annotated RGB-D trajectory dataset. 
%
%
As shown in Fig.~\ref{fig:teaser} and Fig.~\ref{fig:pointcloud}, the point clouds extracted directly from the generated outputs exhibit high spatial accuracy and completeness. 
Such direct 3D reconstruction supports downstream tasks including scene understanding, augmented reality, and robotic perception, underscoring the practical advantages of IDC-Net beyond standard video generation.

\section{Conclusion}
We propose IDC-Net, a unified framework for RGB-D video generation with explicit camera trajectory control. 
%
By jointly modeling RGB and depth within a geometry-aware diffusion paradigm and training on a meticulously annotated metric-aligned RGB-D video sequences dataset, IDC-Net achieves enhanced visual fidelity and robust geometric consistency
%
Comprehensive experiments demonstrate the effectiveness of our approach and highlight its capability to produce metrically accurate outputs that enable direct 3D scene reconstruction.


\bibliography{aaai2026}

\end{document}